\documentclass[10pt,a4paper]{article}
\usepackage[utf8]{inputenc}
\usepackage{amsmath}
\usepackage{array}
\usepackage{tabularx}
\usepackage{amsfonts}
\usepackage{amssymb}
\usepackage{graphicx}
\graphicspath{{./figs/}}
\usepackage[left=2cm,right=2cm,top=2cm,bottom=2cm]{geometry}
\usepackage{authblk}
\usepackage{lineno}
\usepackage{setspace}

\title{Towards automatic detection of wildlife trade using machine vision models}
\author[1]{Ritwik Kulkarni}
\author[1,2,3]{Enrico Di Minin}
\affil[]{\textit {ritwik.kulkarni@helsinki.fi}}
\affil[]{\textit{enrico.di.minin@helsinki.fi}} 
\affil[1]{Helsinki Lab of Interdisciplinary Conservation Science, Department of Geosciences and Geography, FI-00014, University of Helsinki, Finland}
\affil[2]{Helsinki Institute of Sustainability Science (HELSUS), FI-00014, University of Helsinki, Finland}
\affil[3]{School of Life Sciences, University of KwaZulu‐Natal, Durban 4041, South Africa}

\begin{document}

\maketitle
%\linenumbers
%\doublespacing
\begin{abstract}
Unsustainable trade in wildlife is one of the major threats affecting the global biodiversity crisis. An important part of the trade now occurs on the internet, especially on digital marketplaces and social media. Automated methods to identify trade posts are needed as resources for conservation are limited. Here, we developed machine vision models based on Deep Neural Networks with the aim to automatically identify images of exotic pet animals for sale. A new training dataset representing exotic pet animals advertised for sale on the web was generated for this purpose. We trained 24 neural-net models spanning a combination of five different architectures, three methods of training and two types of datasets. Specifically, model generalisation improved after setting a portion of the training images to represent negative features. Models were evaluated on both within and out of distribution data to test wider model applicability. The top performing models achieved an f-score of over 0.95 on within distribution evaluation and between 0.75 to 0.87 on the two out of distribution datasets. Notably, feature visualisation indicated that models performed well in detecting the surrounding context (e.g. a cage) in which an animal was located, therefore helping to automatically detect images of animals in non-natural environments. The proposed methods can help investigate the online wildlife trade, but can also be adapted to study other types of people-nature interactions from digital platforms. Future studies can use these findings to build robust machine learning models and new data collection pipelines for more taxonomic groups. 
\end{abstract}


\section{Introduction}
The unsustainable trade in wildlife (e.g., animals and plants) is a major threat to the conservation of biodiversity globally \cite{Maxwell_et_al_2016} and is increasingly turning to online platforms. Wild populations are hunted or captured at levels that threaten their persistence in order to supply wildlife markets with live animals, plants, and their derived products \cite{'t_Sas-Rolfes_et_al_2019}. Unsustainable trade in wildlife is also often linked to the introduction of exotic species that threaten the persistence of native biodiversity \cite{bellard2016}. In addition, wildlife trade can help facilitate the spread of zoonotic diseases, such as SARS-CoV-2  \cite{BEZERRASANTOS2021}, \cite{DigitalTradeCovidMorcatty2021}. Online markets have created new opportunities to illegally trade wildlife at multiple scales, locally to globally  \cite{DigitalWildlifeTradeLavorgna2014}, \cite{DigitalWildlifeTradeSiriwat2020}. The internet provides cost-effective means for sellers to reach out to active and dormant wildlife users. In turn, this can potentially increase demand for wildlife and their products and incentivize unsustainable hunting and gathering of species in the wild. Innovative solutions are needed to cost-effectively monitor and stop the illegal global online wildlife trade.  

Effective monitoring of global online wildlife trade requires automated content identification   \cite{di2019framework}. In conservation science and practice, the use of machine learning or digital data analysis methods to monitor the online illegal wildlife trade is increasing but is still limited \cite{di2018machine}, \cite{wildlifetradeNLP_Wang_2021}, \cite{wildlifetradeTwitter2019}, \cite{MS_SEEKER}. Compared to other tasks where machine learning methods are being used in conservation science, using machine learning to monitor the online trade in wildlife can be more challenging. Apart from the difficulty of obtaining high-quality labeled training datasets, the partly illegal and hidden nature of the trade, makes data discovery a challenging task\cite{di2019framework}. While natural language processing can be used in cases where text content is structured for analysis or to drive machine learning algorithms \cite{kulkarninews} \cite{NLPStringham2021}, online text content is often unstructured or lacking. Automatic image identification can help identify relevant information from online content when text processing is difficult (e.g non-standard language use or the requirement of language agnostic tools due to multiple languages). However, methods that allow identifying species and products at scale, in the context of wildlife trade, are largely missing. 

In this study, we investigate the application of automatic image classification using deep neural networks on a novel dataset that we constructed and that contains images about the sales of live animals on the internet (see supplementary Fig S5 for a schematic overview of model application). To the best of our knowledge, this is the first attempt at building image classification models to help identify the trade of exotic live animals on the internet. This work differs from other image identification studies in conservation science (e.g., \cite{cameratrap_nature2019}; \cite{cameratrap2018}; \cite{satellite_count2008}; \cite{satellite_agro2015} \cite{satellite_zim2014}; \cite{chen2019wildlife}; \cite{gait2015}), in its application in the wildlife trade context and ability to deduce the context of the target object. The objective here is not to detect a pattern that identifies the animal, but rather to deduce if an animal is present in the image along with identifying the context (e.g., a cage) in which the animal is present. Image classification models can be sensitive to the background information and a change in context adversely affects model performance to detect species as shown in \cite{animal_urbanSingh_2020_WACV}. With no existing datasets available for the task, we created a novel dataset using online resources such as digital market places and crowd sourced databases (see \ref{data_prep}) and obtained images of live animals for sale along with images of animals in the wild. Rather than focusing on identifying individual species, the models aim to differentiate a post showing animals kept in captivity as opposed to posts where animals are found in the wild. This is a necessary condition for the practical applicability of the models in real world scenarios. A total of 24 models were tested, experimenting with five deep learning architectures, two training datasets, three evaluation datasets, three types of training methods and different model sizes (see \ref{meths} for more details). The training objective of the models was to classify between images that have the presence of an animal in natural surroundings in contrast to animals in captive settings. Data were collected from a website trading exotic animals, iNaturalist \footnote{www.inaturalist.org}, Google Image Search \footnote{images.google.com} and Flickr \footnote{www.kaggle.com/hsankesara/flickr-image-dataset}. Models were implemented in PyTorch \cite{pytorchNEURIPS2019_9015} and their performance evaluated for use in real-like conditions of general social media images. Thus, along with evaluating models in a standard manner by splitting training and testing data, we also evaluated the models with \lq out of distribution\rq datasets to measure generalisability, which emphasises the need to go beyond the traditional data splits. Model performance was interpreted using feature visualisation to highlight areas of the image which impact model decision. The best model based on performance metrics and size efficiency was established for this task.     

\section{Methods}\label{meths}
\subsection{Overview}
We trained a total of 24 deep neural network models which spanned a combination of five different architectures, three methods of training, two types of datasets and two sizes of network (large and small for some architectures). The architectures tested were AlexNet \cite{alexnet_krizhevsky2012imagenet}, DenseNet \cite{densenet_huang2017densely}, ResNet \cite{resnet}, Squeezenet \cite{squeezenetiandola2016} and VGG \cite{vgg_simonyan2014very}. We trained a small and large size model for DenseNet (DenseNet121 and DenseNet201), ResNet (ResNet18 and ResNet152) and VGG (VGG11 and VGG19) where the number after each model indicates the number of layers used in that architecture). These models have been highly successful on many image classification tasks and set benchmarks on datasets like Imagenet\cite{Imagent2009} and CIFAR \cite{CIFARkrizhevsky2009learning} while being implemented widely for other applications \cite{model_review_khan2020survey}. Models were trained to discriminate between animals in captivity and the same species in the wild. Thus, the training images consisted of these two classes of data labelled accordingly. As no database was readily available for the task, we created ex novo two datasets for training: i) one that consisted of examples of wild animals (negative class) vs captive animals (positive class) and ii) another one where the negative class also included random images consisting of possible background objects (see \ref{data_prep} for details). The second database was created to investigate the impact of model weights being focused solely on spurious image features as opposed to evaluating the entire context of \lq animal in captivity\rq. Three types of training regimens were used to investigate the effect on model performance (section \ref{modeltraining}): (i) Transfer learning, which entailed using the model pre-trained on the Imagenet data \cite{Imagent2009} and only training the last layer of the network as a binary classifier \cite{transferlearning_pmlr-v27-bengio12a}; (ii) Training from scratch, where the entire model was trained from a blank state for the binary classification task; and (iii) Combination training, in which the model was initially trained as in the transfer learning objective and then later the entire model was trained similar to training from scratch. Training types (ii) and (iii) were restricted to four models consisting of DenseNet and VGG that we found to have best results in type (i). Model performance was determined from accuracy and f-score metrics on three different test sets, namely one within distribution test data and two other test data sets which are out of distribution (see \ref{data_prep}). Evaluating based on out of distribution data helps understand if the models can be applied in a general scenario that usually involves features not present in the training data. As the decision-making of neural networks is not easily interpretable \cite{interpretabilityHuang2020}, we used feature visualisation \cite{featvis2017} for the top two best performing architectures (see \ref{feat_vis}) to gain additional insights on how the model was classifying the images. Feature visualisation enabled us to highlight parts of the image that have the strongest response in calculating the activation of neural units responsible to the classification.

\subsection{Data Preparation}\label{data_prep}
The target images for the positive class needed to be of live mammals (not domestic pets) for sale. We obtained these images from a peer to peer sales website where live mammals are traded and where sellers and buyers interact. As the website contains personal and identifiable information about the sellers, we refrain from publicly mentioning the website and adhere to data privacy regulations (the website was made available to referees during the peer-review of this manuscript). In order to assess the validity of the collected images, we compared them to other images obtained from general social media sites (Twitter, Facebook Marketplace) and Google searches with search queries of the form \lq (animal name) for sale\rq. Upon visual inspection, the images from web searches were found to be identical in nature to the images of animals found on the peer to peer website, and the source of the images appeared to be indistinguishable just by looking at the images. However, it is challenging to obtain data from these sources at volume needed to train models, due to accessibility issues (e.g. automated script blocks in data download), website regulations (e.g. terms prohibiting the use of data) and overall scarcity. See supplementary Fig S4 for example images from various sources. Thus, models trained on the collected data can be widely applicable to multiple digital platforms. We built a customised image downloading script in Python 3.6 (using the library BeautifulSoup) that was executed periodically to scan for new listings on the website and download images along with the associated text for that post. All images for each of the listings were transferred directly onto a cloud storage platform\footnote{www.csc.fi} which hosted all the data for the models. A total of 3051 images covering 58 species (see below) were downloaded over a 4 month period. 

The negative class of images comprised of animals in a non-sale context, essentially wild animals in their natural settings. Image classification models are known to be sensitive to small patterns in the image \cite{featvis2017}. Therefore, to avoid positive detection purely based on particular species characteristics, we obtained images of the same species as in the positive class. We randomly sampled 1000 images from the positive class and made a list of all species mentioned for sale. While the number of new species found diminished strongly as we manually progressed through the images, no new species were found past image number 902 from the sample, therefore indicating that the sample covered most of the species from the positive class. A total of 58 species were found, some of which were included in Appendix I, II and III of CITES (Convention on International Trade in Endangered Species of Wild Fauna and Flora, an international agreement between governments with an aim to ensure that international trade in specimens of wild animals and plants does not threaten the survival of the species) (see Supplementary Table S7 for the full list). Wild images were obtained from the iNaturalist database \footnote{www.inaturalist.org} which is an application allowing wildlife observers to record their sightings and including images. Data were obtained using the iNaturalist API (Application Programming Interface) implemented in Python 3.6 using the library pyinaturalist\footnote{pypi.org/project/pyinaturalist/}. Only images with the quality designated as \lq research grade\rq \hspace{0.5mm} were used and transferred to the same cloud storage as mentioned above. \lq Research grade\rq \hspace{0.5mm} is a designation in the iNaturalist data when an observation passes a quality control process and multiple observers agree on the identification of the species. A total of 2505 images were obtained, thus creating a near balanced class division in the data. The combined 5556 images formed the first dataset designated as \emph{data\_no\_bg} while a second set of data was created with additional images, described below.

As an additional experiment, another set of images was sourced to supplement the negative class of images. The model is intended to assign any image other than an image containing an animal for sale, as a negative detection. However, to test if the model would erroneously base the decisions on random objects in the positive class without taking into account the full context, we made a list of objects present in the positive class images such as table, chair, carpet, cage, human hands etc (see full list in Supplementary Table S8). The list was compiled by manually viewing 500 images of animals in captivity and noting down objects in the background. The images were sourced using Google images as the search engine \footnote{images.google.com} as it allows for specific and flexible search queries such as \lq empty cage for sale\rq. A custom built script was written in Python 3.6 which sourced images based on a search query. Search queries were based on the list of background objects. Overall, 705 images were collected using this method and the combined total of all the images (6261) made the second dataset designated as \emph{data\_bg}. Note that for both datasets the binary class division is near balanced with a slight tilt towards one class (positive in \emph{data\_no\_bg} and negative in \emph{data\_bg}). 

Following common practice, the data were split into 80\% train and 20\% test set. The test set was designated as \emph{test\_in}. However, to gain insights into model performance in real life settings we created two other test sets which contained images from a different source and mimic images from general social media. As the model is intended to monitor the sale of wild animals on digital market places and social media, the assessment of the model in that scenario (images representing general social media) is important. Thus, we had a test dataset that was partly out of distribution (features in the data can be different from the training set). These images were obtained from the Flickr dataset hosted on the website Kaggle\footnote{www.kaggle.com/hsankesara/flickr-image-dataset}. The first out of distribution test data had 998 images from Flickr and 250 from the peer to peer website selling exotic animals which are downloaded separately from the website at a different time not overlapping with the data source used for \emph{test\_in} (total of 1248 images) and was designated as \emph{test\_out\_pet}. This dataset was found to have images from Flickr that has household pets such as cats and dogs. Many wild cat species resemble domestic cats, especially as kittens, while some dogs resemble wolves. Since the models were not trained to detect exact species, it was likely that such images tricked the models into making a positive decision for \lq animal in captivity\rq. Thus, we also made a second out of distribution test set which did not have domestic pets in them to measure the impact this had on model performance. This dataset contained 1246 images (same 250 images sourced from the exotic animal website and 996 from Flickr). Images in the Flickr data have a one line text associated with each of the images which describes the image (e.g. "a man walking a dog"). This facilitate filtering of images based on the description. We used Wordnet\cite{wordnet1995} classes to exclude all images for which the description had a mention of the Wordnet class \lq animal \rq. The Wordnet animal class ensures all dogs and cats are covered along with some other pets that may be in the data (although a visual inspection of a random sample of 100 images from Flickr did not result in animals other than cats and dogs). We called these data \emph{test\_out\_nopet}. See \ref{results} for the impact of the two test sets and the importance to test beyond the 80-20\% split.  

\subsection{Model Training}\label{modeltraining}
Models were trained with three different training protocols. First, all eight models versions (Alexnet, Densenet121, Densenet201, ResNet18, ResNet152, Squeezenet, VGG11 and VGG19) were trained with transfer learning technique on two datasets (\emph{data\_bg and data\_no\_bg}) resulting in 16 models. Model training was carried out using the Pytorch library \cite{pytorchNEURIPS2019_9015} on an NVIDIA GPU (RTX 2070) with a fixed random seed. For transfer learning, models were pre-trained on ImageNet data and only the final layer of the model was modified to have two linear outputs for the two classes. Training was carried out with an early stopping protocol. Thus, training was terminated if there was no improvement in model performance for 15 consecutive epochs. Models with the best performance (lowest Cross-entropy Loss) on the \emph{test\_in} data were saved as final models. Supplementary Fig S1 shows all models tend to approach a plateau after roughly 15 epochs (see supplementary Table S5 for additional information on training parameters). 

We selected four models that were best performing (accuracy above 80\%) among the 16 (Densenet121, Densenet201, VGG11 and VGG19) and carried out two more training experiments. First, we trained these models on the same data as mentioned above but trained the models from scratch without pre-training to see if features for this task could be better captured. Second, we used the combination of transfer learning and training the full model with weights of all the layers modified. For the purpose of combination training, we first trained models for 10 epochs by fine-tuning of final layer but keeping the rest of the weights of the deeper layers frozen, as described above, and then unfroze all layers for the remaining epochs and applied the early stopping protocol. Early stopping was increased to 35 epochs because we had to reduce the learning rate for a stable learning progression (see supplementary Table S5). Similarly, batch size was reduced from 75 to 15 when all layers were trained to avoid memory issues. 

Image augmentation (cropping, flipping, normalising and rotating) was used during training to reduce over-fitting to spurious features and expanding information in the training image distribution \cite{augmentation2019}. Slightly different augmentation pipelines were used for the two classes to remove watermarks in specific places in the two classes, so as to reduce the possibility of the models using watermarks as features (see supplementary Table S6 for details on the augmentation pipeline).   

\subsection{Feature Visualisation}\label{feat_vis}
We used feature visualisation to gain insights on how the model was classifying the images, thus attempting to explore which aspects of an image led to a positive prediction \cite{featvis2017}. A popular method for feature visualisation is to generate saliency maps \cite{saliencySimonyan14deepinside} which highlight the areas of an image that lead to maximum impact on the weight gradients for the target class, thus indicating which human interpretable aspects of the image are involved in decision making by the model. Feature visualisation was implemented using the Flashtorch\footnote{https://github.com/MisaOgura/flashtorch} library. We selected the top ten images from the test set which led to the highest activation of the output unit class, as they indicate the strongest decision bias for identification. We then generated saliency maps for those images. Best representative images are shown in Fig \ref{fig:featvis}, Fig \ref{fig:feat_vis_incrr}, supplementary Fig S2 and Fig S3.

\section{Results} \label{results}
\subsection{Effect of image background on model performance}
\begin{figure}
    \centering
    \includegraphics[scale=0.9]{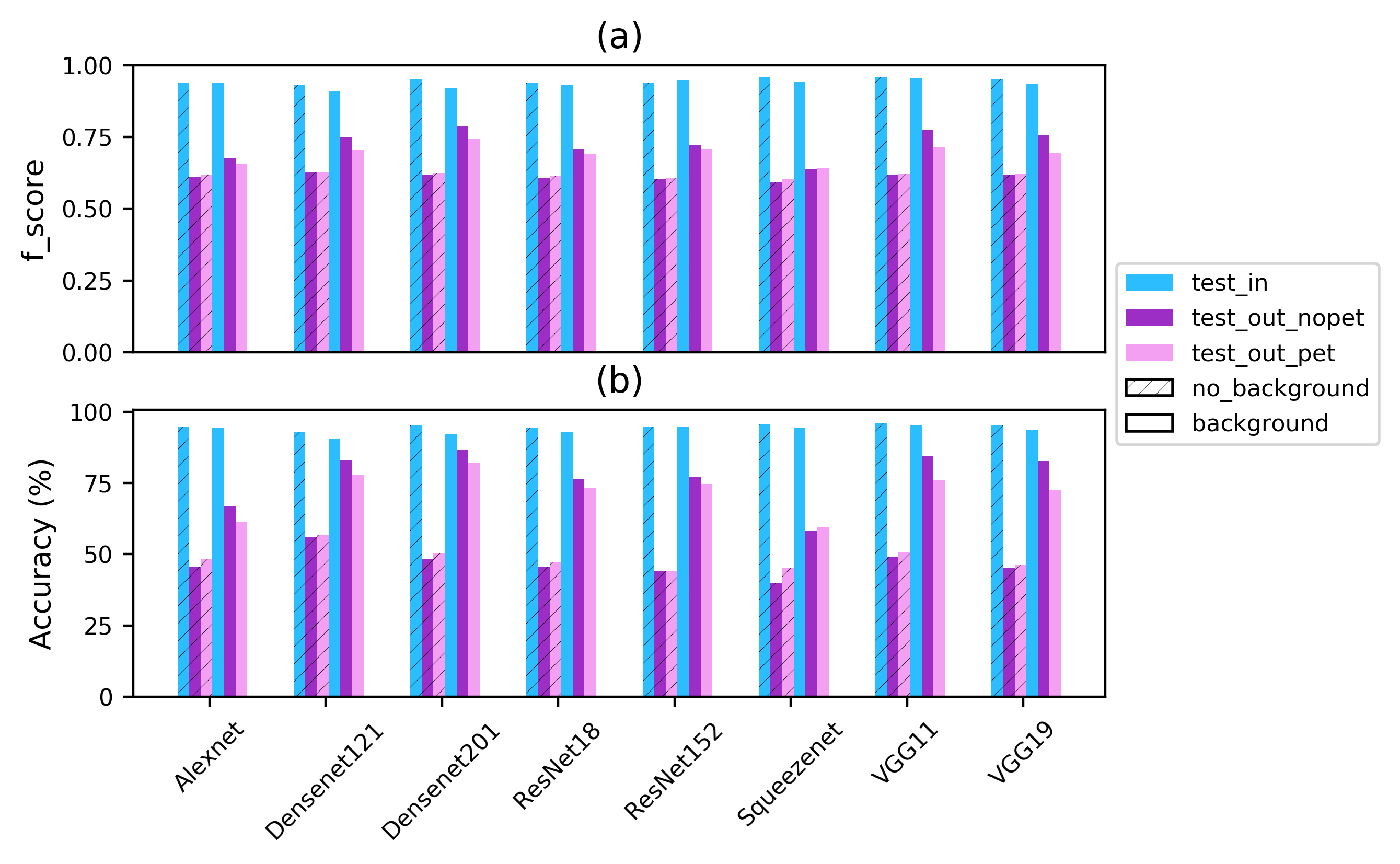}
    \caption{f-score and Accuracy of eight models spanning a combination of two training methods and three test sets. Test methods indicated by \lq background\rq, in which 11\% of training images consisting of random background objects (f-score mean: 0.70, s.d. 0.04; accuracy mean: 74.53, s.d 8.65), and \lq no background\rq (f-score mean: 0.61, s.d. 0.009; accuracy mean: 47.67, s.d. 4.19) for which no specific background images exist. Three test sets indicated by \emph{test\_in}(20\% set aside from training data)(f-score mean: 0.94, s.d.: 0.01; accuracy mean 94.18, s.d.: 1.31), \emph{test\_out\_nopet} (fscore mean: 0.72, s.d.: 0.04; accuracy mean 76.92, s.d.: 9.1) (general images outside training distribution but without any pets) and \emph{test\_out\_pet} (as previous but with images of pets)(fscore mean: 0.69, s.d.: 0.03; accuracy mean 72.92, s.d.: 7.4). See text}
    \label{fig:finetune}
\end{figure}

Models were evaluated using the f-score and accuracy metrics on three different test sets (see \ref{data_prep} for details and supplementary Table S1, Table S2 and Table S3 for all the evaluation figures). Fig \ref{fig:finetune} shows the performance of the eight models trained using two datasets, \emph{data\_bg}, in which a proportion  of training images (11\%)  consisting of random background objects, denoted as \lq background\rq in the figure, and \emph{data\_no\_bg} for which no specific background images exist, denoted as \lq no background\rq. Models were then evaluated on three different test sets \emph{test\_in} (20\% set aside from training data), \emph{test\_out\_nopet} (general images outside training distribution but without any pets) and \emph{test\_out\_pet} (as previous but with some images of pets). Highest performance of all models (f-score mean: 0.94, s.d.: 0.01; accuracy mean 94.18, s.d.: 1.31) was seen on \emph{test\_in} indicating that the models were fairly successful in discriminating the context of the two classes. However, the performance dropped when the models were tested outside their range of training distribution and the top performing models were Densenet121, Densenet201, VGG11 and VGG19 with f-scores mean 0.76, s.d. 0.01, and accuracy mean 84.19, s.d. 1.53. Squeezenet (f-score 0.63, accuracy 58.39) and Alexnet (f-score 0.67, accuracy 66.74) were among the lowest performing models. Overall all models performed remarkably better on the out of distribution datasets when trained with the \emph{data\_bg} dataset compared to \emph{data\_no\_bg} even tough the performance on within distribution test set(\emph{test\_in}) was similar for all models. When tested in the fine tuning paradigm, \emph{test\_bg} had f-score mean: 0.70, s.d. 0.04 and accuracy mean: 74.53, s.d 8.65, while \emph{test\_no\_bg} had f-score mean: 0.61, s.d. 0.009 and accuracy mean: 47.67, s.d. 4.19.  Thus, indicating the added generalisation ability of the model when trained on \emph{data\_bg}, evident only when tested out of distribution. Slightly better performance of all models was found on the \emph{test\_out\_nopet} (f-score mean: 0.72, s.d.: 0.04; accuracy mean 76.92, s.d.: 9.1) compared to \emph{test\_out\_pet} (f-score mean: 0.69, s.d.: 0.03; accuracy mean 72.92, s.d.: 7.4). Images of common pets caused a drop in performance of the models due to a strong similarity of pet images to target positive class animal images. 

\subsection{Effect of training methods}
\begin{figure}
    \centering
    \includegraphics[scale=0.9]{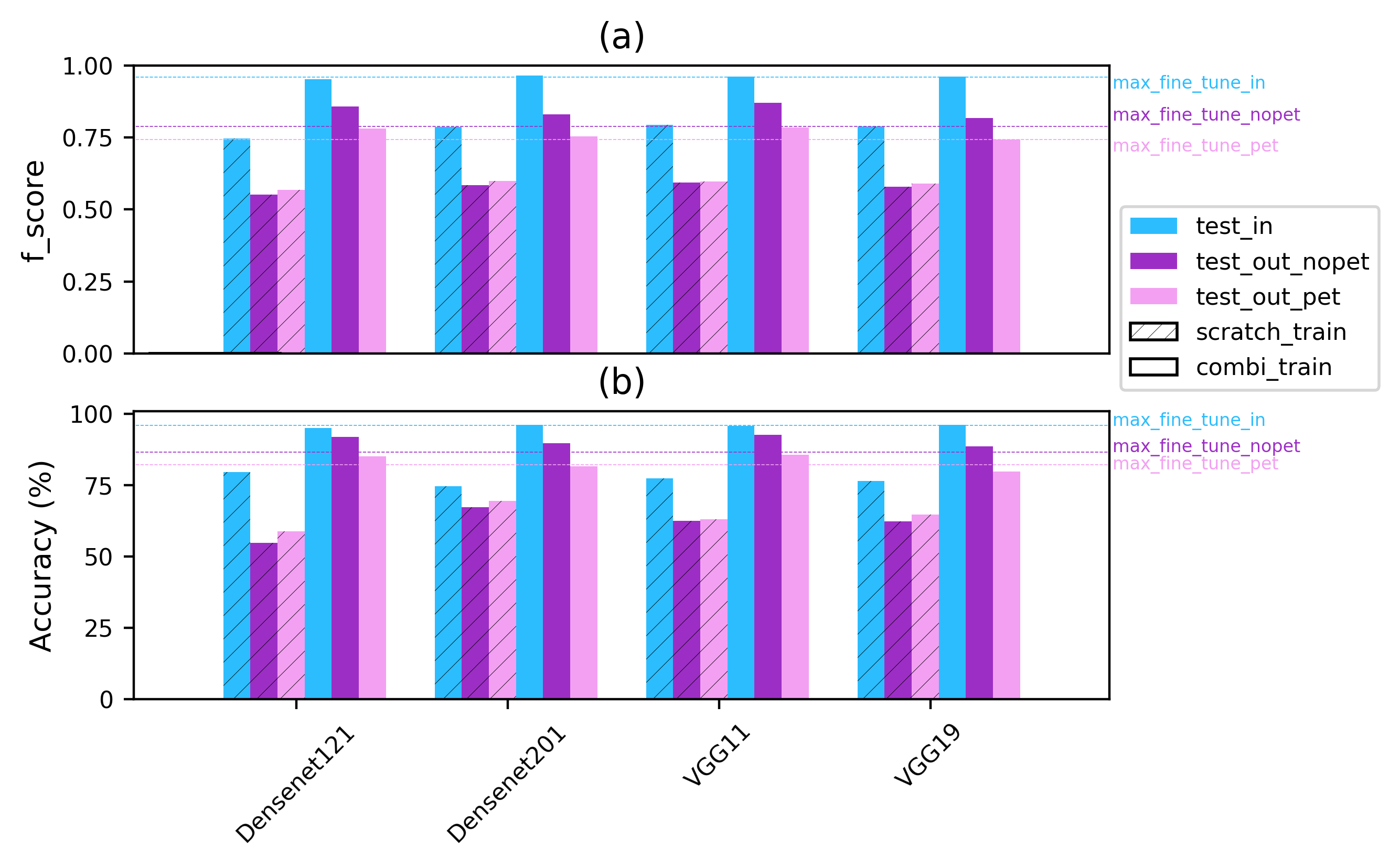}
    \caption{Four top performing models having highest performance in fine-tune paradigm (Densenet121, Densenet201, VGG11 and VGG19) were tested for two additional training paradigms, training the full model from scratch (no pre-training), denoted as \lq scratch\_train\rq (f-score mean: 0.58, s.d.: 0.01; accuracy mean 62.85, s.d.: 4.2),  and combination of fine-tuning and then training the full model once the final layer is semi-trained, denoted as \lq combi\_train\rq (f-score mean: 0.80, s.d.: 0.04; accuracy mean 86.89, s.d.: 4.3). \lq combi\_train\rq training method had higher performance than fine-tune method. Horizontal lines indicate the highest performance in finetune paradigm (Fig \ref{fig:finetune}) for the three respective test sets. Models were trained only with the \emph{data\_bg} dataset due to its advantage on generalisation and evaluated similarly on the three different test sets as for the fine-tuning paradigm.}
    \label{fig:scratch_combi}
\end{figure}

 Accuracy and f-score of the four top performing models (Densenet121, Densenet201, VGG11 and VGG19) is shown in Fig \ref{fig:scratch_combi}. Models trained from scratch consistently under performed (f-score mean: 0.58, s.d.: 0.01; accuracy mean 62.85, s.d.: 4.2) compared to the \lq combi\_train\rq method (f-score mean: 0.80, s.d.: 0.04; accuracy mean 86.89, s.d.: 4.3). Training from scratch was worse in performance than the fine-tuning method, while the \lq combi\_train\rq method performed better than fine-tuning (indicated by horizontal lines in Fig \ref{fig:scratch_combi}). The two top performing models were VGG11 (f-score: 0.87, accuracy 92.69) and Densenet121 (f-score: 0.85, accuracy 91.88), which were the smaller versions of the respective architectures in size. As for the fine-tune train method, evaluation on \emph{test\_out\_nopet} was slightly better than \emph{test\_out\_pet} for \lq combi\_train\rq but comparable for \lq scratch\_train\rq. 

With lower performance on the out of distribution data compared to the within distribution data, we tested the precision of models to identify positive class images only, since this is of practical importance for the application of the models in real life scenarios. Table \ref{tab:prec_pos} lists the precision values for all the models trained with \emph{data\_bg}. The models showed high precision to detect images of animals in captive contexts, except for the \lq scratch\_train\rq case. Highest precision was seen in the \lq combi\_train\rq case, with VGG19 as the top performer. Thus when the model flags the image as positive it has a lower tendency for false positives. Identical precision values across the two datasets are due to the fact that class distribution is near identical in the two test datasets. 

\begin{table}
\centering
\begin{tabular}{ |c||c|c|}
 \hline
 \multicolumn{3}{|c|}{Precision for positive class} \\
 \hline
 Model &  \emph{test\_out\_nopet} & \emph{test\_out\_pet}\\
 \hline
 Alexnet   & 0.95	& 0.94 \\
 Densenet121 &   0.80 &	0.80\\
 Densenet201 & 0.82 &	0.82\\
 ResNet18   & 0.88 &	0.88\\
 ResNet152 & 0.92 &	0.92\\
 Squeezenet & 0.91 & 0.91\\
 VGG11 & 0.92 &	0.92\\
 VGG19 & 0.91 &	0.91 \\
 \hline
 Densenet121 - scratch\_train & 0.63 & 0.63\\
 Densenet201 - scratch\_train & 0.52 & 0.52\\
 VGG11 - scratch\_train & 0.68 & 0.68\\
 VGG19 - scratch\_train & 0.61 & 0.61\\
\hline
 Densenet121 - combi\_train & 0.94 & 0.94\\
 Densenet201 - combi\_train & 0.95 & 0.95\\
 VGG11 - combi\_train & 0.92 & 0.92\\
 VGG19 - combi\_train & \textbf{0.96} & \textbf{0.96}\\
 \hline
\end{tabular}
\caption{Precision to detect positive class only (animals in captivity),  for all models trained with \emph{data\_bg}. VGG19 in the \lq combi\_train\rq paradigm has the top performance on detecting positive hits. }
\label{tab:prec_pos}
\end{table}

\subsection{Model Efficiency}
\begin{figure}
    \centering
    \includegraphics{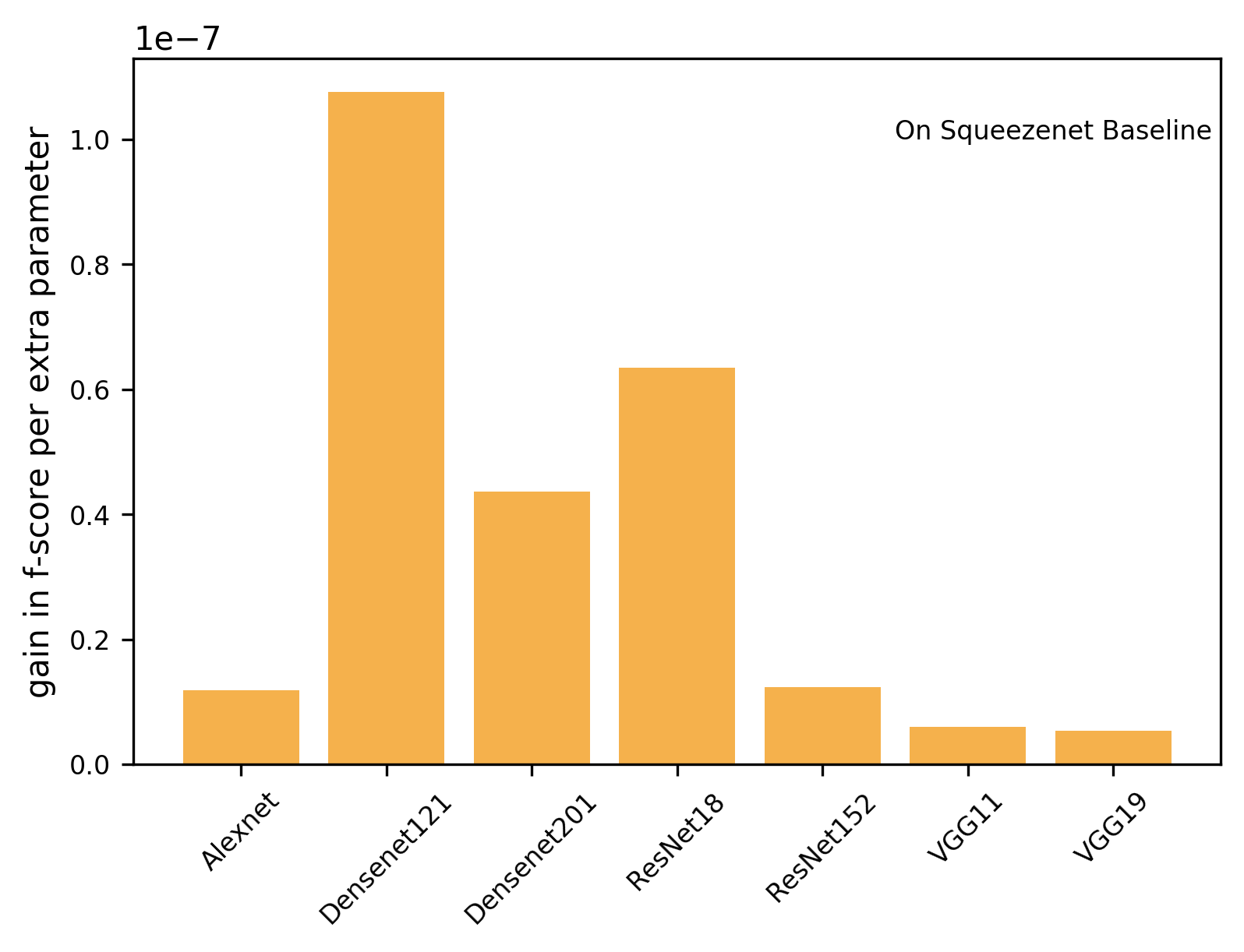}
    \caption{Efficiency of the models measured as the increase in f-score from the Squeezenet baseline, for every additional parameter of the model compared to the number of parameters for Squeezenet. Squeezenet has the lowest number of parameters and also the lowest f-score. Densenet121 even though slightly lower f-score than VGG19 (0.95 vs 0.96) has significantly lower  number of parameters (6,955,906 vs 139,589,442)   }
    \label{fig:gain}
\end{figure}

Fig \ref{fig:gain} shows the gain in performance for every additional parameter added to the architecture as calculated for models trained with the fine-tune paradigm on \emph{test\_in}. Squeezenet was the lowest performing model and also had the lowest number of parameters, thus we used it as the baseline to measure the impact of additional parameters on performance. \lq gain\rq is the ratio of the increase in the f-score of the model for every additional parameter added over the Squeezenet baseline. See supplementary Table S4 for the number of parameters for each model. Densenet121 was found to have the highest gain, indicating that each additional parameter gave the maximum contribution in boosting the f-score above the baseline compared to all other architectures. 

\subsection{Feature Visualisation}
\begin{figure}
    \centering
    \includegraphics[scale=0.31]{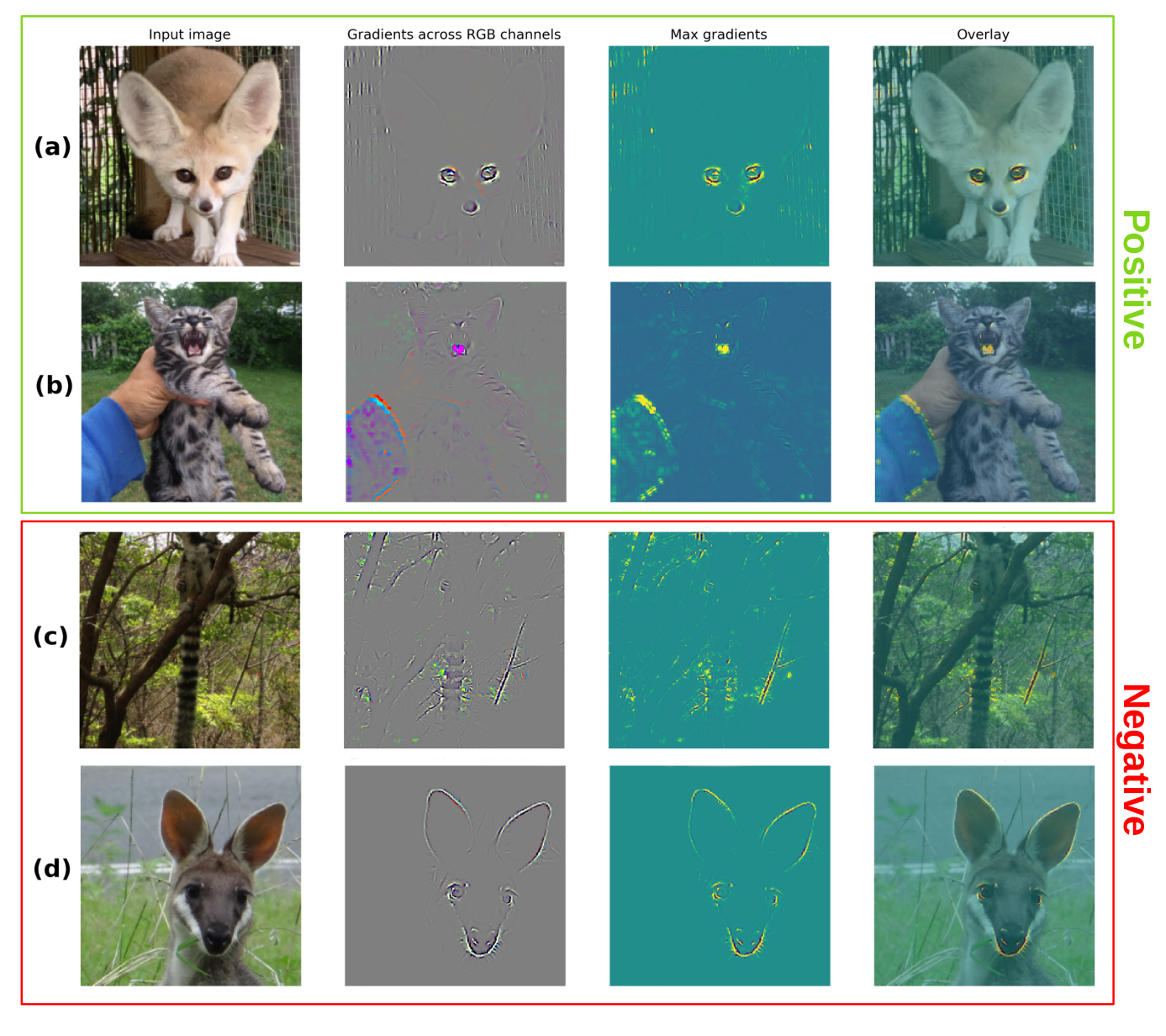}
    \caption{Feature visualisation for images with highest output activation. Top two rows for positive predictions (animals in captive settings) while bottom two rows for negative prediction (animals in natural settings). (a) A captive fennec fox (\textit{Vulpes zerda}) shows strong gradient response for facial features such as eyes and nose along with the background cage; (b) A marbled cat kitten (\textit{Pardofelis marmorata}) evokes strong response for facial features along with a human hand holding it; (c) The tail pattern of a ring-tailed lemur (\textit{Lemur catta}) and the surrounding tree pattern contribute the most to a negative decision; (d) Similar to (a), facial features of a whiptail wallaby(\textit{Macropus parryi}) evoke strong gradient response but the absence of man-made surroundings lead to a negative decision. Overall, feature visualisation exhibits that model decision is based on \lq animal + surroundings\rq . }
    \label{fig:featvis}
\end{figure}

The two top performing models (Densenet121 and VGG11 with  \lq combi\_train\rq) were selected for feature visualisation to observe which features contribute the most for the positive class identification. Images which led to  high output activation for the class labels were selected for visualisation, Fig \ref{fig:featvis} (a) shows a prediction of Densenet121 for a fennec fox (\textit{Vulpes zerda}). Along with the prominent features of the fox, like the eyes and nose, strong gradients are also found for the cage in the background, thus adding context to the animal being in the cage. Similarly, in the prediction for a marbled cat kitten (\textit{Pardofelis marmorata}) (b), the gradients while being strong on the cat shape and fur pattern are also strong on the human hand, again signifying the captive context. Along the same lines images for the negative class, example shown in Fig \ref{fig:featvis} (c), show a strong gradient response for the features of the animal, ring-tailed lemur (\textit{Lemur catta}) and also the surrounding features of the forest, while (d) shows that just the presence of an animal, whiptail wallaby (\textit{Macropus parryi}), is not enough to illicit a positive class response even though it is a close-up of the wallaby similar to captive images. See supplementary Fig S2 for more examples of images of the negative class where human in natural surroundings with no animal presence or humans in urban settings including a close-up of a face, are not enough to have positive class identification. 

\begin{figure}
    \centering
    \includegraphics[scale=0.4]{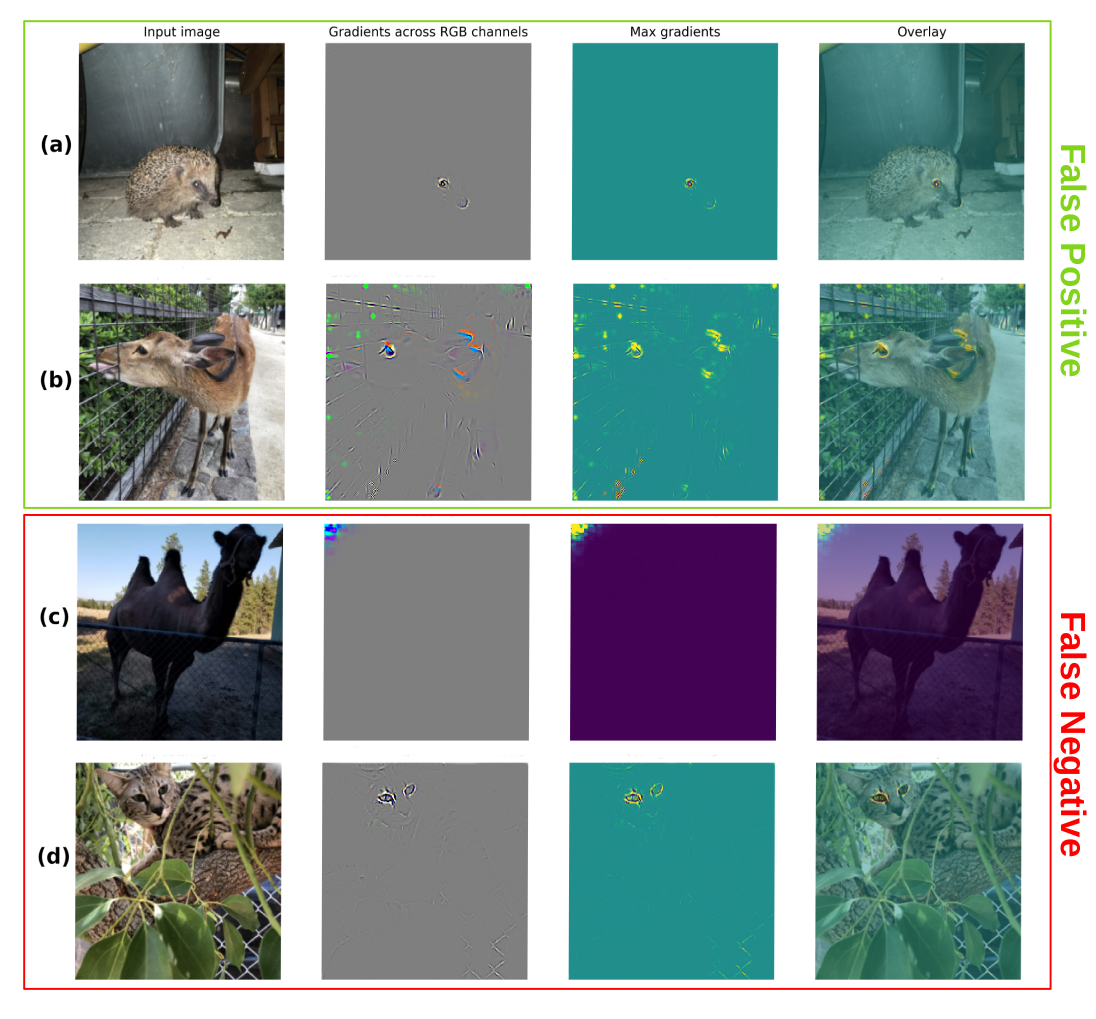}
    \caption{Feature visualisation for incorrect predictions (similar to Fig \ref{fig:featvis}). Top two rows for false positive predictions (animals predicted in captive settings but are wild) while bottom two rows for false negative prediction (animals in predicted natural settings but are captive). (a) European hedgehog (\textit{Erinaceus europaeus}), a wild individual in an urban space, has strong gradient response for facial features but lacks discernible background response; (b) sika deer (\textit{Cervus nippon}), a wild deer along a fence evokes strong gradient response for facial features and the fence which appears like a cage, thus leading to an incorrect response; (c) bactrian camel (\textit{Camelus bactrianus}) a captive individual in a natural looking surrounding, however, the incorrect response is likely due to the model focussing on spurious image features in the top left corner; (d) serval (\textit{Leptailurus serval}), a captive cat in a cage but has dominant natural looking elements like a tree branch and leaves. }
    \label{fig:feat_vis_incrr}
\end{figure}

Similarly, we performed feature visualisation for incorrect predictions of the model for images which produce high activation of output units for the opposite class with respect to the desired class. Fig \ref{fig:feat_vis_incrr} shows False positives on top (a) European hedgehog (\textit{Erinaceus europaeus}), an image where a wild hedgehog is photographed in an urban environment, and (b) sika deer (\textit{Cervus nippon}) again a seemingly wild individual in an urban environment. False negative examples are (c) bactrian camel (\textit{Camelus bactrianus}) for sale, although in this image the camel appears to be in a natural looking background, the error originates from the model weights focusing on an obscure part of the image (sky). While (d) is an image of a serval (\textit{Leptailurus serval}) where even though the cat is in a cage the surroundings are dominated by natural looking elements (leaves and branches).  

\section{Discussion}
In this study, we introduced a method for automatic identification of images pertaining to wildlife trade and evaluated the models by creating a bespoke dataset pertaining to the sale of wild animals. We explained and applied our methods based on images obtained from a website known to post images of mammal species for sale. The novelty of our work is in creating a specialised dataset to distinguish between the context in which an animal is present and demonstrating that the model learns this context as shown by feature visualization for correct prediction of the context. 

Using different models allowed us to learn important aspects useful in the implementation of machine vision applications in detecting the sale of wild animals on the internet. First, it is crucial to evaluate the model on \lq truly\rq unseen data as exhibited in Fig \ref{fig:finetune} and Fig \ref{fig:scratch_combi}. It is common practice to test the models on a small fraction of data set aside from training but derived from the same source as the training data. Unless the training data are of a very large size and covering most of the possible cases that the model will encounter in real applications, there is a high likelihood that models will overfit to the feature set of source \cite{evaluate2013} \cite{overfit2013}. Thus, we recommend that a separate test data is created from a source different to the training data, and that it mimics real world data. Second, it is possible to improve the generalisability of the model and reduce the phenomenon of the model basing decisions on spurious objects by curating the data to include unrelated images containing the random objects that may appear in the target class. As most images of captive exotic animals are in a man made surrounding, including random images of man made objects (negative features) improved the performance of the model in the out of distribution data. This again emphasises the need to evaluate the model on \lq truly\rq unseen data as the difference in performance and improved generalisability was seen on the out of distribution data and not on the 20\% test split. Third, the three different training methods resulted in varying performances of the model. Transfer learning and fine-tuning the final layer for the specified task is an increasingly popular method in many fields including conservation science applications \cite{cameratrap_nature2019} \cite{cameratrap2018}. Training the model from scratch led to poor performance, most likely due to the smaller size of the data than required to train large models. Training deep models requires a large set of data, typically of the order of $10^6$ to capture enough basic features of images. In situations for specific tasks such as wildlife trade, the datasets are quite small and transfer learning can help fill the gap by using pre-trained models. However, even though appreciable  results were seen by fine tuning the final layer only, the best results were obtained using a combination of fine-tuning the last layer till the gradient decent becomes slower, and then letting all the layers learn the weights for subsequent epochs of training. This prevents the deeper layers from being destructively changed during the initial phase of learning and seemed to give added benefits on feature learning for the task when all layers were trained in a slow manner (reducing the learning rate). Fourth, the model is intended to act as an automated service to flag sales of wild animals to inform wildlife trade analyses. At the same time it can be used by online platforms and law enforcement agencies to help prevent the sale of prohibited live animals. In this regard, it is important for the model to not have many false positives which might affect its utility in real applications. The model showed high precision in detecting the images when an animal is present in a captive context. Hence, ensuring that when an image is flagged, it is of a high likelihood to be of interest. Fifth, as Densenet121 had the highest efficiency for the number of parameters used, it might be the best model to be used in such tasks of binary classification of context detection. Further experiments with diverse datasets are required to confirm the efficiency results.

\section{Limitations}
Using feature visualisation to observe areas of images which lead to strong classification decisions, the two top performing models seemed to have learned the general concept of combining animal features with background contexts of captivity (i.e., cages, human body parts, household items, etc). However, there are also some examples (see supplementary Fig S3) when the strongest predictors are some obscure parts of the image. Sometimes, sellers add text to the image (e.g contact information) and the model uses parts of the text as features. When the text is at a consistent location (e.g., name of the website) it has been cropped out during training so the model does not learn the classification based on watermarks, complete removal of spurious text however is not guaranteed. Similarly, text can sometimes be found on the images of wild animals and the impact is mitigated using random crop augmentation (see \ref{modeltraining}). Upon visual inspection of a sample, such cases did not appear to be too frequent in the data. It is also difficult to obtain images particular to the sale of wild animals and the positive class of images are sourced centrally from a single website. The website selected is a peer to peer digital sale platform, ensuring that the sales posts come from a diverse range of sellers and contains a diverse range of animals. Although it is expected that the data represent the nature of digital sales posts in general (examined by manual searched over the web), it is important for future experiments to source images of wild animals for sale from multiple platforms to enhance the model capabilities. Furthermore, Images of domestic pets, common on social media, have the potential to trick the model as seen through the results for \emph{test\_out\_nopet} vs \emph{test\_out\_pet} (Fig \ref{fig:finetune} and Fig \ref{fig:scratch_combi}). Certain dog and cat breeds, especially as cubs, resemble wild canids and felids. Although challenging, training models for species detection as in \cite{cameratrap2018} and \cite{cameratrap_nature2019} will help mitigate the issue to a certain extent or an exclusive pet detection model may be required. From an end user perspective, many times sales posts can be associated with some text information which can be used to further filter and refine relevant posts. In addition, wildlife trade exists for a wide variety of species spanning different taxonomic groups. The data we gathered of animals for sale was restricted by the website for mammals only. Further work is needed to expand the data for different taxonomic groups that would allow a diverse range of stakeholders to apply automated detection of wildlife trade. Finally, while having a strong potential to automatically analyse and flag online sale of threatened species, machine learning methods do possess implementation hurdles as they require specific infrastructure and expert personnel to manage the algorithms. This creates a need for the algorithms to be packaged in easy to use software tools that can be readily implemented with minimal training of on-field stake holders.

In conclusion, together with advanced methods for text classification, our methods offer the potential to investigate online wildlife trade in a cost-efficient manner. We foresee their use as part of scientific research investigating illegal wildlife trade and online monitoring activities by law enforcement agencies and online platforms. The same methods should now be applied to automatically identify wildlife trade images for other taxonomic groups beyond mammals and for multiple online sources at the same time. Future work should now focus on creating new training datasets for further improvement of the models.

\section*{Acknowledgement}
R.K. and E.D.M. thank the European Research Council (ERC) for funding under the European Union's Horizon 2020 research and innovation program (grant agreement 802933).R.K. thanks the KONE Foundation for the research grant. 

\section*{Author Contribution}
R.K. and E.D.M. designed research; R.K. developed the scripts to collect data and  train machine learning models; R.K. and  E.D.M. wrote the paper.

\section*{Data Availability}
Trained models and the source code will be made available post publication at: Etsin (https://etsin.fairdata.fi/)

\section*{Conflict of Interest}
The authors have no conflict of interest to declare.

\bibliographystyle{unsrt}
\bibliography{references}
\end{document}

\maketitle
\beginsupplement
\begin{table}[]
\centering
    \begin{tabular}{|c|c|c|c|c|c|p{1cm}|p{1cm}|}
    \hline
              \textbf{model} & \textbf{data type} &  \textbf{accuracy} &    \textbf{f-score} &      \textbf{precision} &    \textbf{recall} &  \textbf{epoch time (min)} &  \textbf{best epoch} \\
              \hline
            alexnet &    \emph{test\_no\_bg} &     94.86 &  0.94 &  0.94 &  0.94 &      5.56 &          20 \\
            alexnet &    \emph{test\_bg} &     94.40 &  0.94 &  0.94 &  0.94 &      6.5 &          12 \\
        densenet121 &    \emph{test\_no\_bg} &     93.05 &  0.93 &  0.93 &  0.93 &      6.08 &          45 \\
        densenet121 &    \emph{test\_bg} &     90.64 &  0.91 &  0.90 &  0.90 &      7.2 &           5 \\
        densenet201 &    \emph{test\_no\_bg} &     95.31 &  0.95 &  0.95 &  0.95 &      6.5 &          50 \\
        densenet201 &    \emph{test\_bg} &     92.24 &  0.92 &  0.91 &  0.92 &      6.99 &          30 \\
           resnet18 &    \emph{test\_no\_bg} &     94.22 &  0.94 &  0.94 &  0.93 &      7.71 &          29 \\
           resnet18 &    \emph{test\_bg} &     93.04 &  0.93 &  0.928 &  0.929 &      7.8 &          33 \\
          resnet152 &    \emph{test\_no\_bg} &     94.58 &  0.94 &  0.944 &  0.941 &      6.35 &          29 \\
          resnet152 &    \emph{test\_bg} &     94.80 &  0.948 &  0.946 &  0.943 &      6.72 &          89 \\
         squeezenet &    \emph{test\_no\_bg} &     95.67 &  0.957 &  0.956 &  0.956 &      5.81 &          56 \\
         squeezenet &    \emph{test\_bg} &     94.32 &  0.944 &  0.94 &  0.942 &      6.59 &          29 \\
            vgg11bn &    \emph{test\_no\_bg} &     95.94 &  0.96 &  0.959 &  0.959 &      6.32 &          81 \\
            vgg11bn &    \emph{test\_bg} &     95.20 &  0.954 &  0.951 &  0.952 &      7.13 &          37 \\
            vgg19bn &    \emph{test\_no\_bg} &     95.22 &  0.952 &  0.951 &  0.951 &      5.88 &          52 \\
            vgg19bn &    \emph{test\_bg} &     93.52 &  0.936 &  0.935 &  0.935 &      8.65 &          36 \\
 densenet121scratch &    \emph{test\_bg} &     79.62 &  0.745 &  0.767 &  0.740 &      7.19 &           3 \\
 densenet201scratch &    \emph{test\_bg} &     74.56 &  0.786 &  0.742 &  0.722 &      7.5 &           6 \\
 vgg11bnscratch &    \emph{test\_bg} &     77.36 &  0.793 &  0.772 &  0.757 &      7.03 &           6 \\
     vgg19bnscratch &    \emph{test\_bg} &     76.40 &  0.787 &  0.764 &  0.747 &      7.89 &           6 \\
   densenet121combi &    \emph{test\_bg} &     94.96 &  0.952 &  0.943 &  0.942 &      6.67 &          16 \\
   densenet201combi &    \emph{test\_bg} &     96.16 &  0.964 &  0.962 &  0.960 &      7.13 &          16 \\
       vgg11bncombi &    \emph{test\_bg} &     95.84 &  0.962 &  0.955 &  0.955 &      8.85 &          17 \\
       vgg19bncombi &    \emph{test\_bg} &     96.08 &  0.961 &  0.960 &  0.958 &      6.91 &          12 \\
       \hline
\end{tabular}
    \caption{Performance of all twenty four models evaluated on \emph{test\_in} (20\% split from training data). \emph{test\_bg} - when models were trained on data in which a poportion of images consisted of random background objects. \emph{test\_no\_bg} - as previous but without images of random background objects. See main text. \lq best epoch\rq is the epoch which has highest f-score on validation set while \lq epoch time\rq is the time to complete the best epoch. }
    \label{tab:my_label}
\end{table}

\begin{table}[]
    \centering
    \begin{tabular}{|c|c|c|c|c|c|p{1cm}|}
    \hline
             \textbf{model} & \textbf{data type} &  \textbf{accuracy} &    \textbf{f-score} &      \textbf{precision} &    \textbf{recall} &  \textbf{eval time (min)} \\ 
    \hline
            alexnet &    \emph{test\_no\_bg} &     45.70 &    0.61 &  0.64 &    0.45 &      0.94 \\
            alexnet &    \emph{test\_bg} &     66.74 &    0.68 &  0.77 &    0.64 &      0.95 \\   
        densenet121 &    \emph{test\_no\_bg} &     56.14 &    0.63 &  0.69 &    0.54 &      0.90 \\            
        densenet121 &    \emph{test\_bg} &     82.97 &    0.75 &  0.82 &    0.77 &      0.80 \\            
        densenet201 &    \emph{test\_no\_bg} &     48.27 &    0.62 &  0.66 &    0.48 &      0.94 \\            
        densenet201 &    \emph{test\_bg} &     86.58 &    0.79 &  0.85 &    0.81 &      0.82 \\            
           resnet18 &    \emph{test\_no\_bg} &     45.38 &    0.61 &  0.64 &    0.45 &      0.94 \\            
           resnet18 &    \emph{test\_bg} &     76.38 &    0.71 &  0.81 &    0.72 &      0.88 \\            
          resnet152 &    \emph{test\_no\_bg} &     44.01 &    0.61 &  0.63 &    0.44 &      0.94 \\            
          resnet152 &    \emph{test\_bg} &     77.10 &    0.72 &  0.83 &    0.73 &      0.92 \\            
         squeezenet &    \emph{test\_no\_bg} &     40.00 &    0.59 &  0.60 &    0.40 &      0.93 \\            
         squeezenet &    \emph{test\_bg} &     58.39 &    0.64 &  0.71 &    0.56 &      0.91 \\            
            vgg11bn &    \emph{test\_no\_bg} &     48.99 &    0.62 &  0.66 &    0.48 &      0.94 \\           
            vgg11bn &    \emph{test\_bg} &     84.49 &    0.77 &  0.88 &    0.80 &      0.92 \\           
            vgg19bn &    \emph{test\_no\_bg} &     45.30 &    0.62 &  0.64 &    0.45 &      0.96 \\           
            vgg19bn &    \emph{test\_bg} &     82.73 &    0.76 &  0.86 &    0.78 &      0.91 \\            
 densenet121scratch &    \emph{test\_bg} &     54.85 &    0.55 &  0.58 &    0.51 &      0.64 \\            
 densenet201scratch &    \emph{test\_bg} &     67.22 &    0.58 &  0.62 &    0.58 &      0.53 \\            
     vgg11bnscratch &    \emph{test\_bg} &     62.40 &    0.59 &  0.65 &    0.57 &      0.68 \\            
     vgg19bnscratch &    \emph{test\_bg} &     62.32 &    0.58 &  0.62 &    0.56 &      0.62 \\            
   densenet121combi &    \emph{test\_bg} &     91.88 &    0.86 &  0.93 &    0.89 &      0.93 \\            
   densenet201combi &    \emph{test\_bg} &     89.79 &    0.83 &  0.92 &    0.86 &      0.96 \\
   vgg11bncombi &    \emph{test\_bg} &     92.69 &    0.87 &  0.93 &    0.89 &      0.94 \\            
   vgg19bncombi &    \emph{test\_bg} &     88.59 &    0.82 &  0.91 &    0.85 &      0.95 \\            
\hline
\end{tabular}
    \caption{Performance of all twenty four models evaluated on \emph{test\_out\_pet} (out of distribution data which also contains images of domestic pets). \emph{test\_bg} - when models were trained on data in which a poportion of images consisted of random background objects. \emph{test\_no\_bg} - as previous but without images of random background objects. See main text. \lq eval time\rq is the time to complete the evaluation on the full test dataset. }
    \label{tab:my_label}
\end{table}

\begin{table}[]
    \centering
    \begin{tabular}{|c|c|c|c|c|c|p{1cm}|}
    \hline
              \textbf{model} & \textbf{data type} &  \textbf{accuracy} &    \textbf{f-score} &      \textbf{precision} &    \textbf{recall} &  \textbf{epoch time (min)} \\ 
    \hline
            alexnet &    \emph{test\_no\_bg} &     48.27 &    0.62 &  0.66 &    0.48 &      0.94 \\
            alexnet &    \emph{test\_bg} &     61.26 &    0.66 &  0.74 &    0.59 &      0.95 \\
        densenet121 &    \emph{test\_no\_bg} &     56.77 &    0.63 &  0.69 &    0.55 &      0.90 \\
        densenet121 &    \emph{test\_bg} &     78.02 &    0.71 &  0.79 &    0.72 &      0.80 \\
        densenet201 &    \emph{test\_no\_bg} &     50.44 &    0.62 &  0.67 &    0.49 &      0.94 \\
        densenet201 &    \emph{test\_bg} &     82.19 &    0.74 &  0.82 &    0.76 &      0.82 \\
           resnet18 &    \emph{test\_no\_bg} &     47.39 &    0.61 &  0.65 &    0.47 &      0.94 \\
           resnet18 &    \emph{test\_bg} &     73.21 &    0.69 &  0.79 &    0.69 &      0.89 \\
          resnet152 &    \emph{test\_no\_bg} &     44.10 &    0.61 &  0.63 &    0.44 &      0.94 \\
          resnet152 &    \emph{test\_bg} &     74.65 &    0.71 &  0.81 &    0.71 &      0.93 \\
         squeezenet &    \emph{test\_no\_bg} &     45.06 &    0.61 &  0.63 &    0.45 &      0.93 \\
         squeezenet &    \emph{test\_bg} &     59.34 &    0.64 &  0.71 &    0.57 &      0.92 \\
            vgg11bn &    \emph{test\_no\_bg} &     50.52 &    0.62 &  0.67 &    0.50 &      0.94 \\
            vgg11bn &    \emph{test\_bg} &     75.86 &    0.71 &  0.82 &    0.72 &      0.93 \\
            vgg19bn &    \emph{test\_no\_bg} &     46.43 &    0.62 &  0.65 &    0.46 &      0.96 \\
            vgg19bn &    \emph{test\_bg} &     72.65 &    0.69 &  0.80 &    0.69 &      0.92 \\
 densenet121scratch &    \emph{test\_bg} &     58.78 &    0.57 &  0.61 &    0.54 &      0.64 \\
 densenet201scratch &    \emph{test\_bg} &     69.52 &    0.60 &  0.63 &    0.60 &      0.53 \\
     vgg11bnscratch &    \emph{test\_bg} &     63.11 &    0.60 &  0.65 &    0.58 &      0.68 \\
     vgg19bnscratch &    \emph{test\_bg} &     64.63 &    0.59 &  0.63 &    0.58 &      0.62 \\
     densenet121combi &    \emph{test\_bg} &     85.08 &    0.78 &  0.88 &    0.81 &      0.94 \\
   densenet201combi &    \emph{test\_bg} &     81.63 &    0.75 &  0.87 &    0.77 &      0.95 \\
       vgg11bncombi &    \emph{test\_bg} &     85.72 &    0.79 &  0.88 &    0.81 &      0.93 \\
       vgg19bncombi &    \emph{test\_bg} &     79.79 &    0.74 &  0.86 &    0.76 &      0.96 \\
     \hline
\end{tabular}
    \caption{Performance of all twenty four models evaluated on \emph{test\_out\_nopet} (out of distribution data which does not contain images of domestic pets). \emph{test\_bg} - when models were trained on data in which a proportion of images consisted of random background objects. \emph{test\_no\_bg} - as previous but without images of random background objects. See main text. \lq eval time\rq is the time to complete the evaluation on the full test dataset. }
    \label{tab:my_label}
\end{table}
\begin{table}[]
    \centering
    
    \begin{tabular}{|c|c|}
    \hline
       \textbf{Model} &   \textbf{Number of Parameters} \\
       \hline
     Alexnet &   57012034 \\
     Densenet121 &    6955906 \\
     densenet201 &   18096770 \\
     ResNet18 &   11177538 \\
     ResNet152 &   58147906 \\
     Squeezenet &     736450 \\
     VGG11 &  128780034 \\
     VGG19 &  139589442 \\
     \hline
    \end{tabular}
    \caption{Number of trainable parameters for each of the architectures used}
    \label{tab:my_label}
\end{table}

\begin{figure}
    \centering
    \includegraphics[scale=0.75]{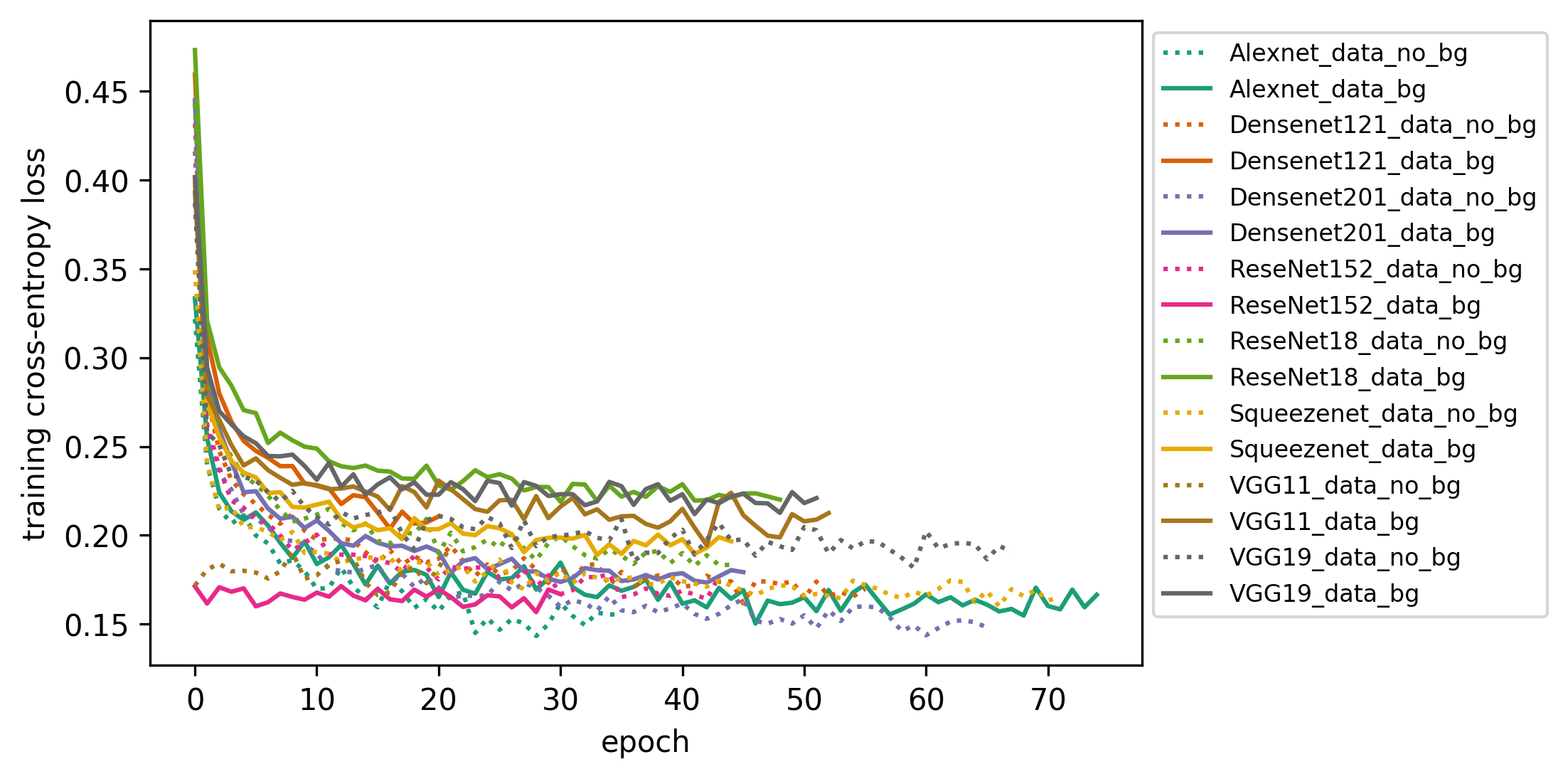}
    \caption{Training history of models in the fine-tune paradigm. Cross-entropy loss against successive training epochs. Early stopping protocol was implemented. See main text.  }
    \label{fig:trainhist}
\end{figure}

\begin{figure}
    \centering
    \includegraphics[scale=0.4]{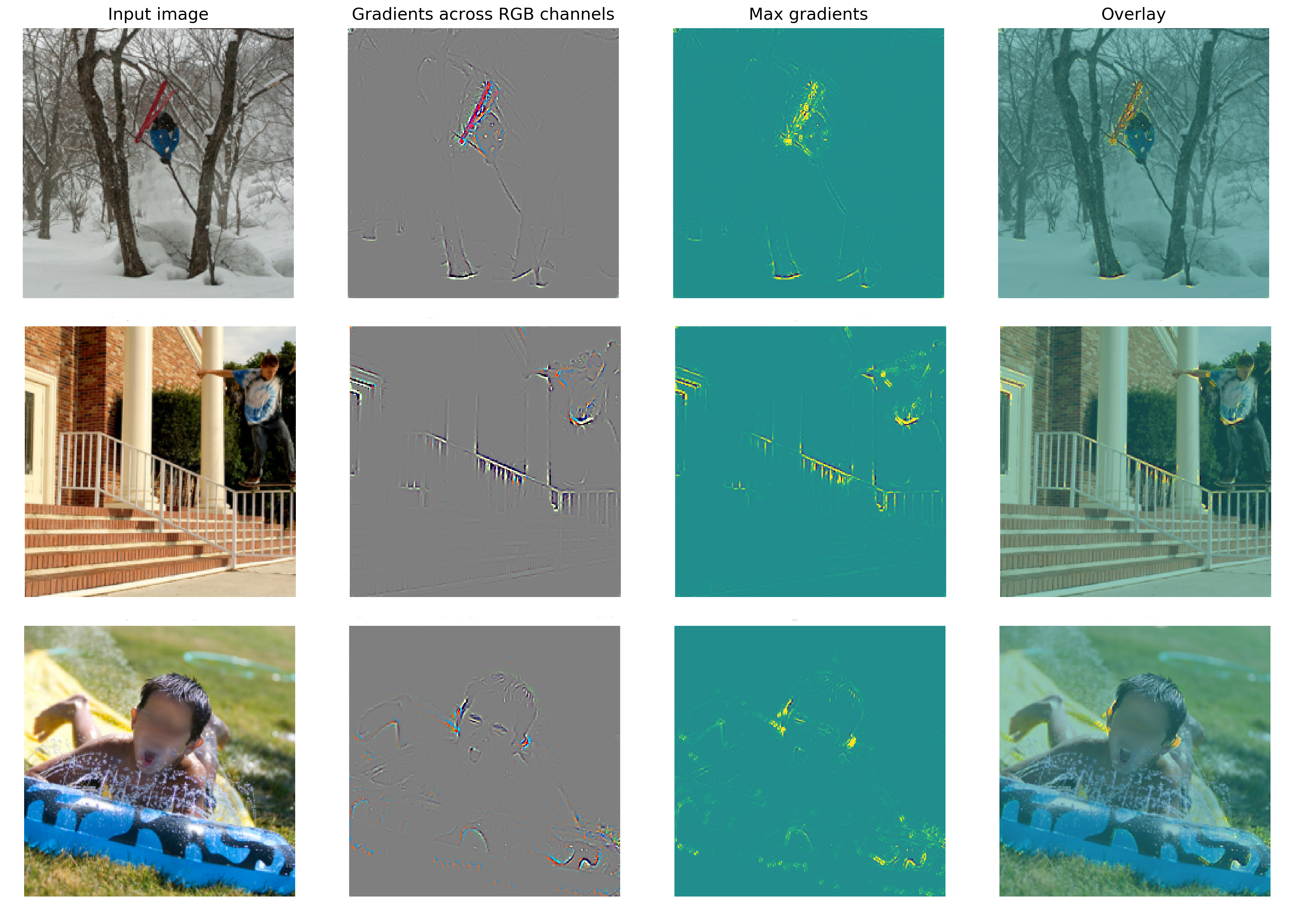}
    \caption{Feature visualisation to illustrate model decision making on negative predictions. See main text. Top row - image with dominant natural features but absence of an animal. Middle row - urban environment as in positive class images but an absence of an animal leads to negative decision. Bottom row - facial features and urban environment, a dominant trend in positive class, fails illicit a positive prediction due to the absence of an animal (face blurred for data privacy). }
    \label{fig:feat_wild}
\end{figure}

\begin{figure}
    \centering
    \includegraphics[scale=0.4]{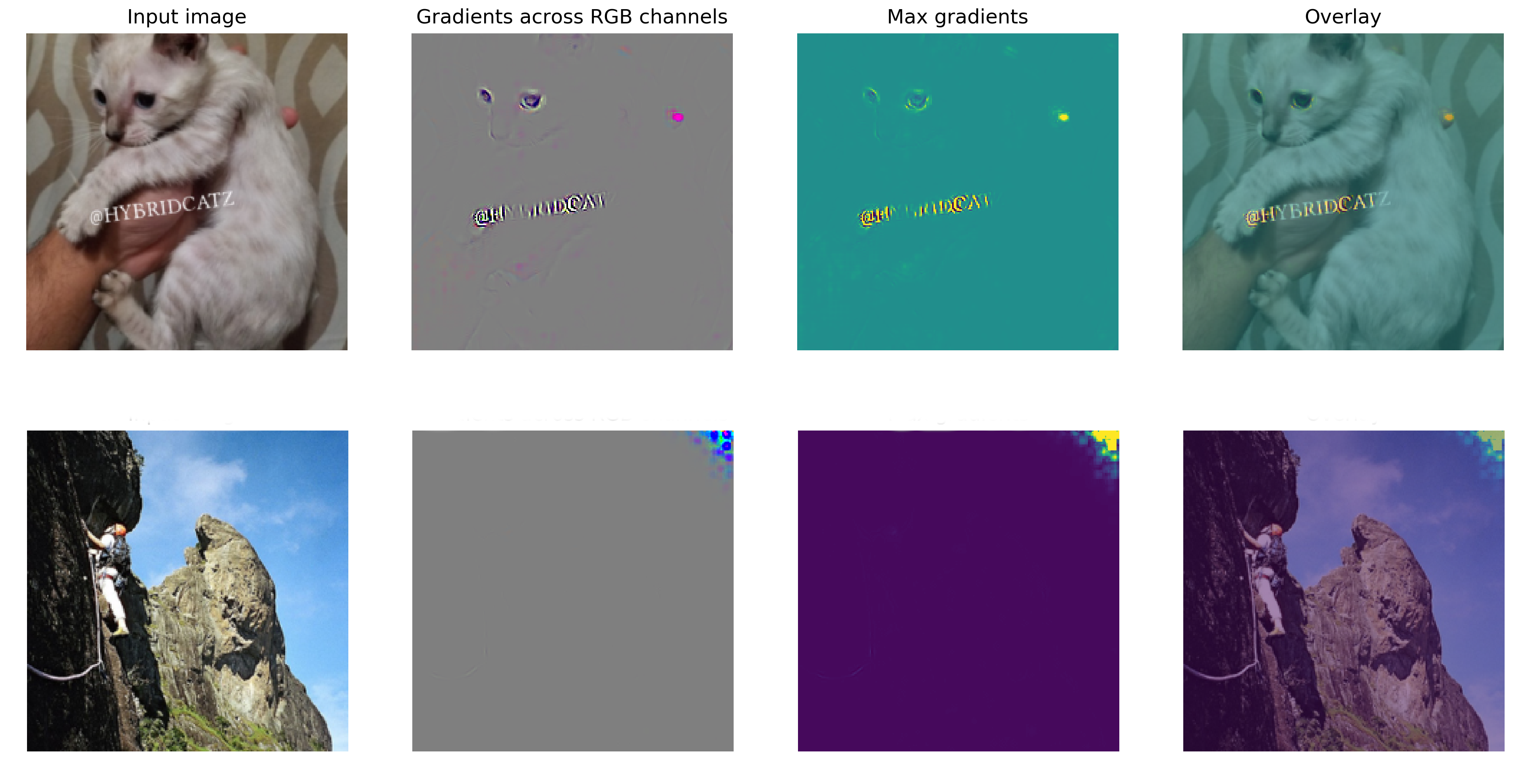}
    \caption{Feature visualisation of model in cases when decision is based on spurious features. Top row - along with facial features, the text in the image produces a strong gradient response, thus overfitting on spurious features. Bottom row - strong gradient are concentrated on the top right corner of the image }
    \label{fig:feat_err}
\end{figure}

\begin{table}[]
    \centering
    \begin{tabular}{|c|c|}
    \hline
        Parameter  &  Value \\
    \hline
    batch size     &  75 (fine tune final layer); 15 (all layers) \\
    epoch early stopping threshold & 15 (fine tune final layer); 35 (all layers) \\
    learning rate  &  0.001 ((fine tune final layer); 0.0005 (all layers) \\
    momentum  &  0.9 \\
    loss function  &  binary cross-entropy loss \\
    optimiser  &  stochastic gradient descent \\
    \hline
    \end{tabular}
    \caption{Hyperparameters of the model. batch size is reduced from 75 to 15 when all layers are trained due to high memory requirement of training additional parameters. Early stopping threshold is increased from 15 to 35 when all layers are trained due to different learning dynamics for the two training paradigms. Similarly learning rate is reduced when all layers are trained to avoid unstable learning behaviour due to large gradients.}
    \label{tab:my_label}
\end{table}

\begin{table}[]
    \centering
    \begin{tabular}{|c|}
         \hline
         \textbf{Image Augmentation Pipeline} \\
         \hline
          Resize (224,224)\\
          Normalize([0.485, 0.456, 0.406], [0.229, 0.224, 0.225]) \\
          Center Crop (0.85) -Only for positive class (removes text on most images)\\
          Random Crop (0.8) -Helps remove miscellaneous text\\
          Random Crop (0.8) on 30\% random sample \\
          Horizontal Flip on 50\% random sample \\
          $180^{\circ}$ Rotation on 50\% random sample \\
          Resize (224,224)\\
          \hline
    \end{tabular}
    \caption{Sequence and parameters for image augmentation during model training. During prediction, only Resize and Normalise are implemented. }
    \label{tab:my_label}
\end{table}

\begin{table}[]
    \centering
    \begin{tabular}{|l|l|l|l|}
\hline
                Scientific Name & CITES Appendix &                 Scientific Name & CITES Appendix \\
\hline
        Acomys cahirinus &            NA &           Glaucomys volans &              NA \\
          Aonyx cinereus &            NA &  Hydrochoerus hydrochaeris &              NA \\
       Aotus trivirgatus &             II &           Hystrix cristata &              NA \\
     Arctictis binturong &            III &                 Lama glama &              NA \\
        Ateles geoffroyi &           I/II &                Lemur catta &                I \\
             Bison bison &            NA &         Leptailurus serval &               II \\
     Bradypus variegatus &             II &          Lontra canadensis &               II \\
         Bubalus bubalis &            NA &                  Lynx lynx &               II \\
      Callithrix jacchus &             II &                 Lynx rufus &               II \\
     Caluromys philander &            NA &            Macropus parryi &              NA \\
      Camelus bactrianus &            NA &             Macropus rufus &              NA \\
     Camelus dromedarius &            NA &      Mustela putorius furo &              NA \\
           Canis latrans &            NA &               Nasua narica &              III \\
             Canis lupus &           I/II &               Nasua narica &              III \\
         Caracal caracal &           I/II &              Octodon degus &              NA \\
          Cebus imitator &            NA &       Pardofelis marmorata &                I \\
          Cervus elaphus &    I/II/III/NC &         Petaurus breviceps &              NA \\
           Cervus nippon &            NA &          Phalanger mimicus &               II \\
     Chinchilla lanigera &              I &               Potos flavus &              III \\
    Chlorocebus aethiops &             II &   Prionailurus bengalensis &             I/II \\
 Chlorocebus pygerythrus &             II &    Prionailurus viverrinus &               II \\
    Crossarchus obscurus &            NA &              Procyon lotor &              NA \\
                 Cynomys &            NA &             Saguinus midas &               II \\
    Didelphis virginiana &            NA &           Saguinus oedipus &                I \\
    Dolichotis patagonum &            NA &           Saimiri sciureus &               II \\
            Equus quagga &            NA &      Sciurus variegatoides &              NA \\
     Erinaceus europaeus &            NA &      Tolypeutes tricinctus &              NA \\
         Felis margarita &             II &              Vulpes vulpes &           III/NC \\
         Genetta genetta &            NA &               Vulpes zerda &               II \\
\hline
\end{tabular}
    \caption{List of all the species manually collected from positive class data and used to obtain negative class data from the iNaturalist dataset. See main text. CITES appendix is mentioned if the species is listed in CITES data}
    \label{tab:my_label}
\end{table}

\begin{table}[]
    \centering
    \begin{tabular}{|c|}
    \hline
        Google Images Search Query for Background Objects   \\
         \hline
         shirt tops \\
hands \\
house flooring \\
house walls \\
carboard boxes in house \\
table in house \\
wood \\
empty cage for sale \\
couch in house \\
 window in house \\
backyard fence \\
bedding in house \\
towel in house \\
soft sofa throws in house \\
basket in house \\
hay \\
grass \\
simple bathroom tiles \\
plant in house \\
indoor curtain in house \\
bedroom door in house \\
tree in backyard \\
 \hline
    \end{tabular}
    \caption{Search queries to obtain images from Google Image Search for random background objects. The list was complied by manually observing a sample of positive class images and noting down objects in the background}
    \label{tab:my_label}
\end{table}

\begin{figure}
    \includegraphics[scale=0.25]{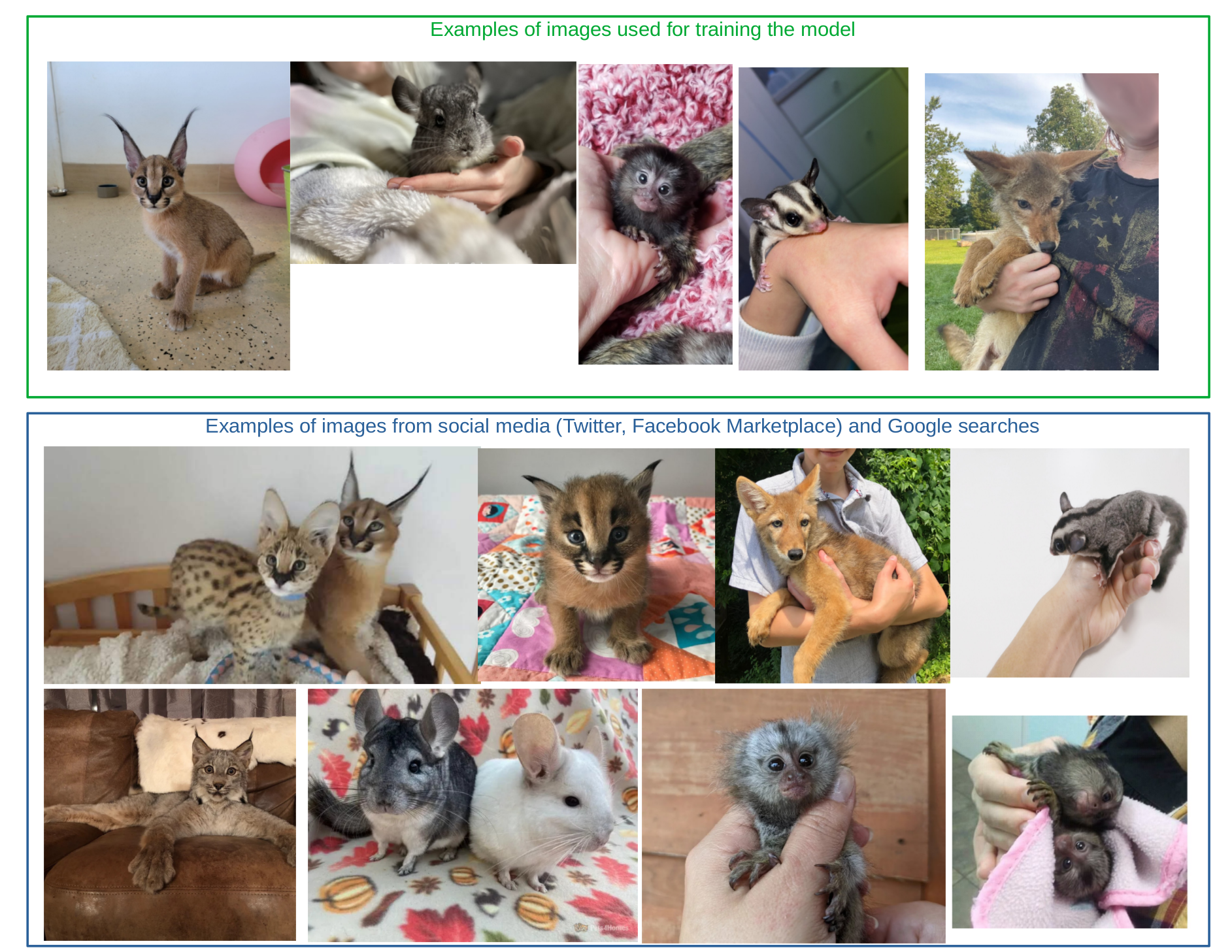}
    \caption{Visual comparison of the nature of images obtained from the website for sale of exotic animals, used to train models (top panel), and the nature of images found on the general web  such as social media and digital markets (bottom panel). Examples illustrate that the source of images is not distinguishable by just observing the images, thus making the results of the model more widely applicable beyond the website.}
    \label{fig:my_label}
\end{figure}

\begin{figure}
    \centering
    \includegraphics[scale=0.17]{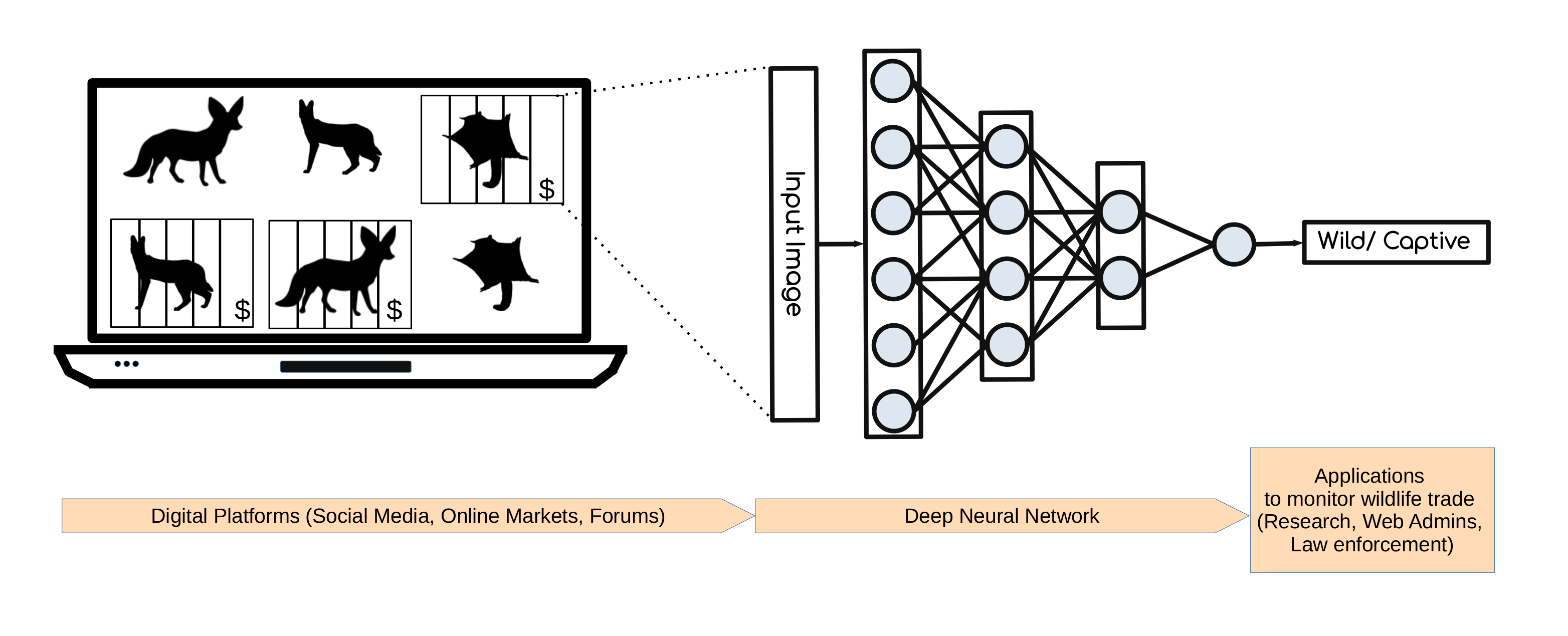}
    \caption{Schematic overview of the intended framework to detect and monitor the sale of potentially illegal wildlife over the internet. Image classification models analyse images on the web (social media, forums, marketplaces) and flag images that maybe selling a prohibited species. This tool can then be used by stake holders such as, researches to study the prevalence of wildlife trade, web admins to take down posts that violate the law, or law enforces to investigate illegal trade related aspects}
    \label{fig:my_label}
\end{figure}